\pdfoutput=1

\documentclass[11pt]{article}

\usepackage[]{acl}

\usepackage{times}
\usepackage{latexsym}
\usepackage{multirow}
\usepackage{helvet} 
\usepackage{courier}  
\usepackage{graphicx}
\usepackage[T1]{fontenc}
\usepackage[utf8]{inputenc}
\usepackage{microtype}
\usepackage{subfigure}
\usepackage{booktabs}

\usepackage{amsmath}
\usepackage{xspace}
\usepackage{bbm}
\usepackage{amsthm}
\usepackage{amsfonts,amssymb}

\title{A Multibias-mitigated and Sentiment Knowledge Enriched Transformer for Debiasing in Multimodal Conversational Emotion Recognition}

\author{Jinglin Wang\textsuperscript{1}\Thanks{ Jinlin Wang and Fang Ma contribute equally to this work and share the co-first authorship.}, Fang Ma\textsuperscript{1}\footnotemark[1], Yazhou Zhang\textsuperscript{2,3}, Dawei Song\textsuperscript{1}\Thanks{ Dawei Song is the corresponding author.} \\
   \textsuperscript{1}Beijing Institute of Technology\\
   \textsuperscript{2}Zhengzhou University of Light Industry\\
   \textsuperscript{3}State Key Lab. for Novel Software Technology, Nanjing University\\
   \texttt{\{jinglinwang,mfang,dwsong\}@bit.edu.cn} \\
   \texttt{yzhou\_zhang@bit.edu.cn}
}

\date{}

\begin{document}

\maketitle
\begin{abstract}
Multimodal emotion recognition in conversations (mERC) is an active research topic in natural language processing (NLP), which aims to predict human's emotional states in communications of multiple modalities, e,g., natural language and facial gestures. Innumerable implicit prejudices and preconceptions fill human language and conversations, leading to the question of whether the current data-driven mERC approaches produce a biased error. For example, such approaches may offer higher emotional scores on the utterances by females than males. In addition, the existing debias models mainly focus on gender or race, where multibias mitigation is still an unexplored task in mERC. In this work, we take the first step to solve these issues by proposing a series of approaches to mitigate five typical kinds of bias in textual utterances (i.e., gender, age, race, religion and LGBTQ+) and visual representations (i.e, gender and age), followed by a \textbf{M}ultibias-\textbf{M}itigated and sentiment \textbf{K}nowledge \textbf{E}nriched bi-modal \textbf{T}ransformer (MMKET). Comprehensive experimental results show the effectiveness of the proposed model and prove that the debias operation has a great impact on the classification performance for mERC. We hope our study will benefit the development of bias mitigation in mERC and related emotion studies.

\end{abstract}

\section{Introduction} 

\begin{figure}[ht]
     \resizebox{0.48\textwidth}{!}{
  \includegraphics[width=\linewidth]{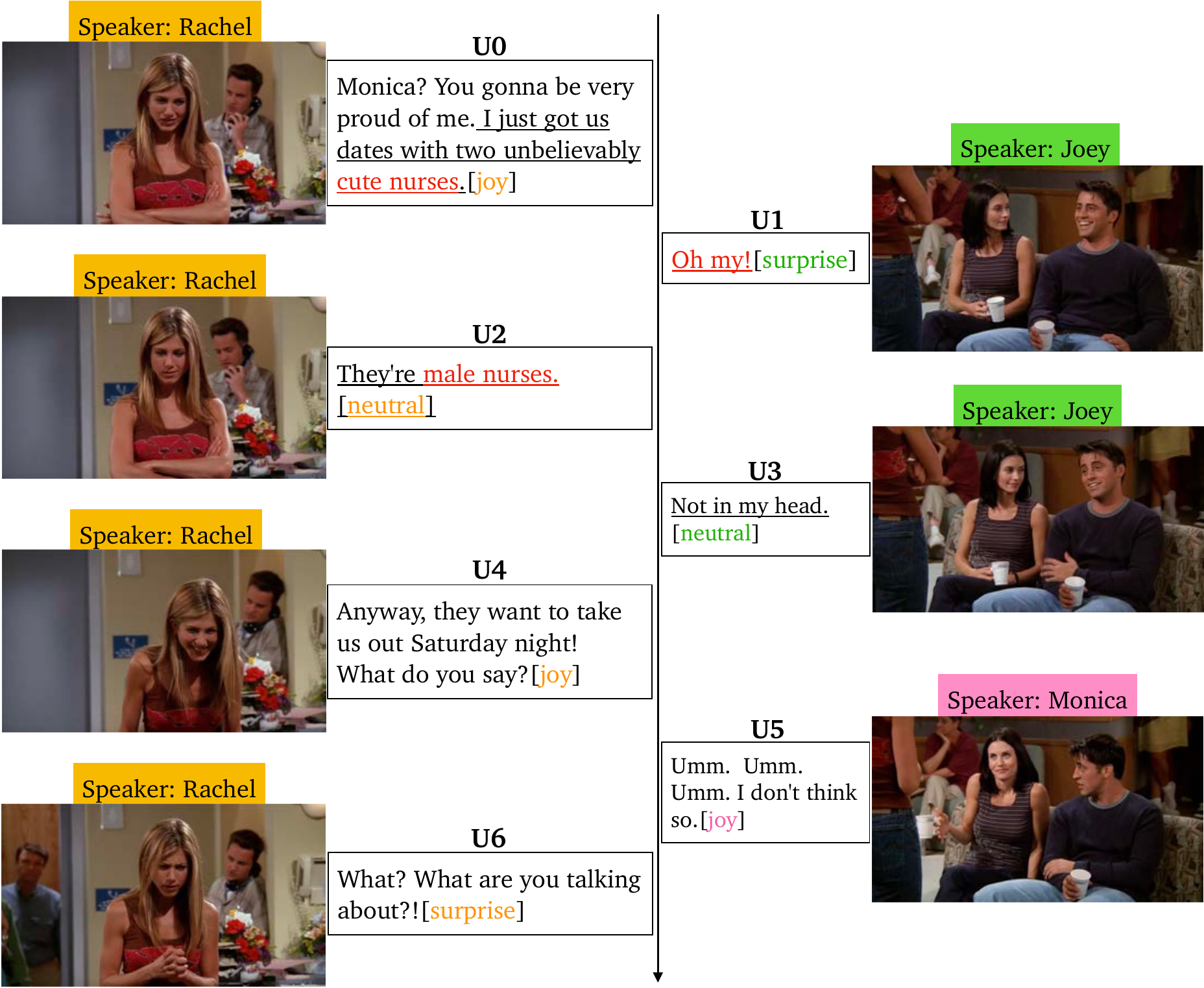}
  }
  \caption{An illustrative example of gender bias in the multimodal dialogue dataset (MELD). There are three speakers in this conversation: Rachel, Joey, and Monica. In the bracket at the end of each utterance is the emotion type (e.g., joy, surprise, etc.) of it. The \underline{underlined} sentences and the words highlighted in \textcolor{red}{red} express an apparent gender bias, and such bias is further amplified by Joey's facial gestures.}
  \label{gender_bias_MELD}  
\end{figure}

Whether people realize them or not, innumerable implicit prejudices and preconceptions fill human language, and are conveyed in almost all data sources, such as news, reviews and conversations~\cite{misra2016seeing}. Such prejudices are known to hurt specific groups, and infringe their rights. For example, these two utterances, ``older people are not interested in digital technology'', or ``women are pleasant to look slim'', reveal the age and gender biases.

Recent research has shown that pre-trained word representations, e.g., word embeddings (in which each word is represented as a vector in the semantic space), tend to amplify the bias in the data~\cite{kurita2019measuring,webster2020measuring}. The male names have been proved more likely to be associated with career-related terms than female names, by calculating the similarity between their embeddings~\cite{caliskan2017semantics}. African-American names are also shown to be more likely to be associated with unpleasant terms than European-American names~\cite{nadeem2020stereoset}. Unconsciously learning such implicit biases from a dataset that is sampled from all kinds of data sources, leads to the fact that the learned models may further amplify the harmful bias (such as gender or race) when they make decisions~\cite{goyal2019making,srinivasan2021worst}. In addition, the biased error will propagate to downstream tasks. For example, coreference resolution systems exhibit a gender bias due to the use of biased word embeddings~\cite{rudinger2018gender}. Facial recognition applications have also been proved to perform worse for the inter-sectional group ``darker females'' than for either darker individuals or females~\cite{buolamwini2018gender}.

In view that human language is multi-modal in nature, human bias also exists in multimodal conversations, e.g., textual and visual utterances. Figure~\ref{gender_bias_MELD} shows an example of the gender bias in a multimodal dialogue dataset \footnote{Trigger Warning: This paper contains examples of biases and stereotypes seen in society and language representations. These examples may be potentially triggering and offensive. These examples are meant to bring light to and mitigate these biases, and it is not an endorsement.}. Joey makes an association with a beautiful female nurse and expresses a significant smile when Rachel says ``cute nurse'', but when Rachel says ``they are male nurses'', he shows a disappointed-looking facial expression, although his textual response seems neutral. 

Therefore, human bias naturally resides in the multimodal expression of emotions in conversations. 

There has been a great body of literature in debiasing for computer vision~\cite{buolamwini2018gender} and pre-trained language models~\cite{wang2020double}. However, the existing debias models mainly focus on only one kind of prejudice, e.g., gender or race, where multibias mitigation is still an unexplored task in mERC. This leaves us with a research question: \textit{Whether the current data-driven multi-modal emotion recognition in conversations approaches produce a biased error or not?}  

To answer this question, we first propose a series of approaches for debiasing multiple types of bias in multimodal (i.e., textual and visual) conversations. For textual utterances, we propose mitigating five types of bias, including gender, age, race, religion, and LGBTQ+ in word embedding. 
 
For visual utterances, we first propose a subspace-projection-based debiasing approach to mitigate two typical visual biases, i.e., gender and age. It constructs a subspace for each type of visual bias and identifies the type of bias in the visual representation by projecting the representation into the corresponding subspace.

To incorporate the proposed multimodal debiasing methods into the mERC task that involves conversational context modeling, cross-modality interactions capturing and the use of sentiment knowledge, we propose a \textbf{M}uiltibiases \textbf{M}itigated and sentiment \textbf{K}nowledge \textbf{E}nriched \textbf{T}ransformer (MMKET) as a unified framework. Specifically, it is a bimodal Transformer involving a contextual attention layer to capture the contextual interactions, a bimodal cross-attention layer to capture the cross-modal interactions, and a sentiment attention layer to enrich the debiased representation with sentiment knowledge. 

Empirical evaluation has been carried out on two benchmark datasets, and the experimental results shows that the proposed multimodal debiasing methods can effectively mitigate the corresponding biases. We also prove that debiasing the representation of multimodal utterances has a remarkable impact on the performance of mERC models.

    






\begin{figure*}[ht]
  \centering
    \resizebox{1\textwidth}{!}{
  \includegraphics[width=\linewidth]{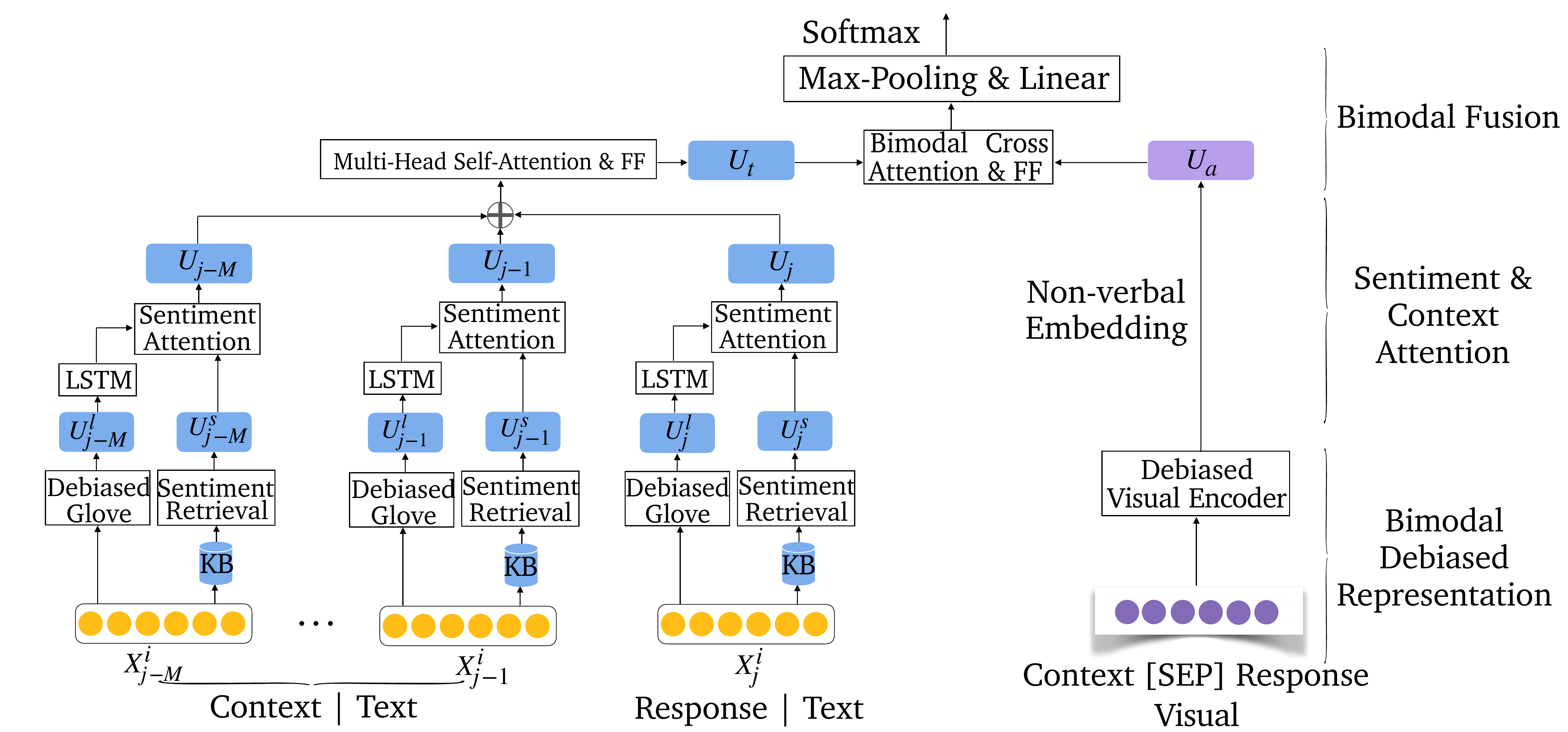}
  }
  \caption{Overall architecture of our proposed Multibias-Mitigated and sentiment Knowledge enriched Transformer model.}
  \label{modelarchitecture}  
\end{figure*}

\begin{table*}[ht]
\centering
\begin{tabular}{cc}
\hline
\textbf{Bias Type} & \textbf{Word Pairs}\\
\hline
                                          
Gender    & \begin{tabular}[l]{@{}c@{}}woman-man, girl-boy, she-he, mother-father, daughter-son, gal-guy, female-male \end{tabular}                                                                   \\
Race      & \begin{tabular}[l]{@{}c@{}}slave-secondary, group-tribe, easy task-cake walk, master-primary \end{tabular}             \\
Age       & \begin{tabular}[l]{@{}c@{}}young-old, health-disease, work-retirement, education-pension \end{tabular}           \\
Religion  & \begin{tabular}[l]{@{}c@{}}Christian-Muslim, Christianity-Islam,  Christ-Allah, Jesus-Muhammad\end{tabular}                                           \\
LGBTQ+    & \begin{tabular}[l]{@{}c@{}}homosexuals-they, husband/wife-spouse, dad/father/mom/mother-parent\\ 
\end{tabular} \\ \hline
\end{tabular}
\caption{\label{word_pairs}
The five pre-defined sets of word pairs, where the main difference between each pair of words captures the corresponding bias.  
}

\end{table*}

\section{Generation of Bias}
Models and algorithms have never independently created bias. Social bias is exhibited in multiple components of a NLP system, including the training corpus, pre-trained models (e.g., word embeddings), and algorithms themselves~\cite{caliskan2017semantics,garg2018word}.

\paragraph{\textbf{Datasets: The Soil of Bias.}}
The dataset is the basis of model training. The bias in the dataset comes from the unbalanced dataset samples and biased labels. For example, gender bias manifests itself in training data that features more examples of men than women, an unbalanced dataset. In the process of label annotation, the annotators will transfer personal bias to the data, where the algorithm absorbs, thus produces a biased model. Sometimes, such bias is due to a lack of domain expertise~\cite{plank2014learning} or preconceived notions and stereotypes held by the annotators~\cite{sap2019risk}.

\paragraph{\textbf{Bias in Word Embeddings.}}
Word embeddings are often trained from large and human-created corpora that contain multifarious biased raw data. Recent literature has demonstrated that gender bias is encoded in word embeddings~\cite{may2019measuring}. For example, Bolukbasi highlights that ``programmer'' is more closely associated with ``man'' while ``homemaker'' is more closely associated with ``woman'' in word2vec embeddings trained on the Google News dataset~\cite{bolukbasi2016man}. And word embeddings connect medical doctors more frequently to male pronouns than female pronouns~\cite{caliskan2017semantics}. Furthermore, pre-trained word embeddings are often used without access to the original data. Social bias in word embeddings will propagate to downstream tasks, which can further amplify social bias. Studies show that machine translation systems tend to link occupations to their stereotypical gender, e.g., linking ``doctor'' to ``he'' and ``nurse'' to ``she''~\cite{prates2019assessing}. 

\section {Debiasing Methods}

\subsection{Mitigating Multiple Biases in GloVe}
The recent debiasing models~\cite{bolukbasi2016man,wang2020double} have only focused on removing gender bias in pre-trained word embeddings, particularly GloVe~\cite{pennington2014glove}, which has surfaced several social biases~\cite{spliethover2021bias}. In this paper, we propose to mitigate five types of biases in GloVe embeddings, i.e., \texttt{gender}, \texttt{race}, \texttt{religion}, \texttt{age}, and \texttt{LGBTQ+}. Methodologically, we extend the existing Double-Hard Debias method, to multiple types of bias. 

\paragraph{\textbf{Hard Debias~\cite{bolukbasi2016man}.}}
Hard Debias is a commonly adopted debiasing strategy in NLP. It projects a pre-trained word embedding vector into a subspace orthogonal to an inferred bias subspace (i.e., direction of a particular type of bias), which is constructed based on a set of pre-defined word pairs (e.g., \textit{young} vs. \textit{old}) characterizing the bias. 

To extend it to multiple types of bias mitigation, we manually define a set of $n$ characterizing word pairs for each type of bias based on typical data biases. Table \ref{word_pairs} shows a range of representative examples.

\paragraph{\textbf{Double Hard Debias~\cite{wang2020double}.}}
Wang et al. discovered that word frequency twists the bias direction, and proposed the Double-Hard Debias method.To find an intermediate subspace that can mitigate the effect of word frequency on the bias direction, Wang uses the clustering accuracy of highly biased words as an indicator to iteratively test the principal components of the word embedding space. 

Specifically, the Double Hard Debias method includes the following steps (taking age bias for example):
\par(a)  Let $W$ be the vocabulary of the word embeddings we aim to debias. Pick the top biased young and elderly words $W_y, W_e \in W$, according to the Cosine similarity of their embeddings to the age direction computed earlier.
\par(b) Calculate the principal components of $W$ (measured by their projections onto the age direction) as the candidate frequency direction. Repeat steps (c)-(e) for each candidate dimension $u_i$ respectively.
\par(c) The top biased word embeddings are mapped to an intermediate space orthogonal to $u_i$ to mitigate the frequency-related bias: ${w}_{bias}^{\prime} = {w}_{bias} - (u_{i}^{T}w_{bias})u_i$, where $w_{bias} \in W_y, W_e$.
\par(d) Apply the Hard Debias method. The characterizing word pairs $D_{t_1},D_{t_2},...D_{t_n} \subset W$  are used here to substract the bias projection from the top biased word embeddings: $\hat{w}_{bias} = \operatorname{Hard} \operatorname{Debias}\left(w_{bias}^{\prime}\right)$. The detailed steps of $\operatorname{Hard} \operatorname{Debias}$ can be found in the original paper.
\par(e) Cluster the $\hat{w}_{bias}$, and then compute the corresponding accuracy and append to $S_{debias}$.
 
The purpose of debiasing is to make the top biased words (e.g., words about young and elderly) less separable. So the lower clustering accuracy in $S_{debias}$, the better debiasing effect that removing $u_i$ has (i.e. the top biased words are mixed up). In other words, we filter out the $u_i$ that causes the most significant decrease in the clustering accuracy and then remove it. Let $j = \arg \min_{i}  S_{debias}$, we get the frequency-debiased word embeddings : ${w}^{\prime} = {w} - (u_{j}^{T}w)u_j$, where $w \in W$. Then, apply the Hard Debias method to ${w}^{\prime} $ to obtain the output age-debiased word embedding: $\hat{w} = \operatorname{Hard} \operatorname{Debias}\left({w}^{\prime}\right)$.

The algorithm operates on the five types of bias sequentially, i.e., the debiased word embedding of the first type serves as the input for the second, and so forth. Finally, we get the multibias-mitigated pre-trained word embeddings, which can be used in our proposed MMKET model.

\subsection{Mitigating Multiple Biases in Visual Representation} 

Recent research shows that gender and age bias accounts for a noticeable portion of visual bias~\cite{drozdowski2020demographic}. To mitigate them, we propose two methods: Visual Hard Debias and Projection Debias methods. The Visual Hard Debias method can mitigate the superficial gender and age bias in visual representation. Then we devise the Projection Debias method to further mitigate finer-grained visual bias.

\paragraph{\textbf{Visual Hard Debias.}}
We assume that, for each type of visual bias, there is a pre-defined set of $n$ image pairs $V_1, V_2,...,V_n \in S$ (e.g., male-female or young-old), which represent the bias. The images are selected randomly from IMDB-WIKI~\cite{Rothe-IJCV-2018}, a publicly available face image dataset with gender and age labels. Let $\vec{v}$ denote an image's visual representation. Let
$u_i =  {\textstyle \sum_{p \in V_i }^{}\vec{p}/|V_i|  }$ be the mean of the image representations of $V_i$ in the pre-defined image set. The visual bias subspace $VB$ is spanned by the first $k (\ge 1)$ eigen-vectors of $VC)$, by applying Singular Value Decomposition (SVD) on it.
\begin{equation}
    VC := \sum_{i=1}^{m} \sum_{v \in V_i}^{} ( \vec{v} - u_i )^T(\vec{v}-u_i)/|V_i|
 \label{visual_bias_subspace}
\end{equation}
 Here the $k$ is set to 1. As a result, the bias subspace $VB$ becomes a bias direction $\overrightarrow{VB}$.
After getting the visual bias subspace, each image representation $\vec{v}$ is debiased through: $\tilde{v}=\vec{v}-(\overrightarrow{VB}^{T} \cdot \vec{v} ) \overrightarrow{VB}$.

\paragraph{\textbf{Projection Debias.}}
In order to further mitigate the finer-grained gender and age bias in the image representation, we propose a new visual debias method, namely Projection Debias. Specifically, it projects the image representation twice into the bias subspaces  (e.g., \textit{male} vs. \textit{female}, \textit{young} vs. \textit{old}) respectively. 
\begin{figure}[ht]
  \centering
  \resizebox{0.48\textwidth}{!}{
  \includegraphics[width=\linewidth]{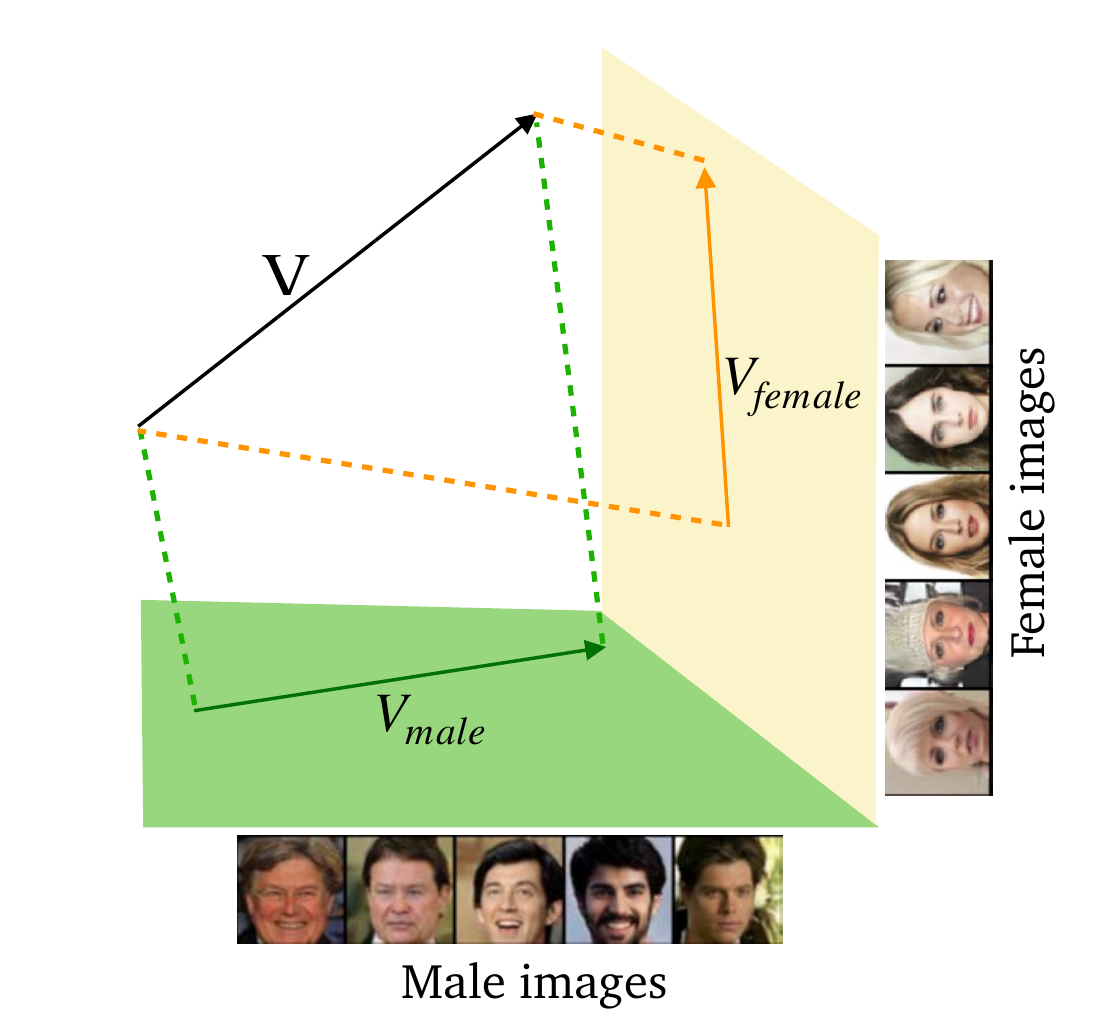}
  }
  \caption{Projection Debias.} 
  \label{projection_debias}  
\end{figure}
By subtracting the two projections from the original visual representation, we get the final debiased representation. Figure \ref{projection_debias} shows the Projection Debias method on gender bias. 

First we use the IMDB-WIKI to define four sets of images $U_{i}$, where $ i \in [1, 2, 3, 4]$, corresponding to the female, male, young and old respectively. The we compute the bias subspace as: 
    
\begin{equation}
  {B}_{i}={\vec{u}_{i}}^\mathrm{T} \otimes \vec{u}_{i}
 \label{projection_bias_subspace_male}
\end{equation}
where $ i \in [1, 2, 3, 4]$,  $\vec{u}_{i}$ is the first principal component of $U_{i}$ computed through Principal Component Analysis ( $\operatorname{PCA})$. 
$\mathrm{T}$ means the transpose operation, and  $\otimes$ means outer product operation. Then we can get the corresponding projection-debiased visual representation through:

\begin{equation}
  \hat{v}=\tilde{v} - \sum_{i=1}^{4} ({B}_{i} \times \tilde{v})
 \label{debiased_projection_res}
\end{equation}

Then we get the two-step debiased visual representation $\hat{v}$ for an image.

\section{The Proposed MMKET Model}
We outline the Multibias-mitigated and Sentiment Knowledge Enriched Transformer (MMKET) model (c.f. Figure \ref{modelarchitecture}). In the proposed framework, we apply the Transformer~\cite{vaswani2017attention} to leverage the debiased contextual and multimodal (text and visual) clues to predict the emotions of the target utterance, due to its ability to capture the context and fast computation. The main ideas are: (1) a multi-modal encoder to create textual and visual representations of contexts and responses, including debiased word embedding (GloVe) and debiased visual representation from the pre-trained EfficientNet Network~\cite{tan2019efficientnet}. (2) The text representation is enriched by sentiment knowledge. (3) The context-aware attention mechanism is proposed to effectively incorporate conversational context. (4) The text representation and non-verbal embedding are forwarded through a self-attention layer and a feed-forward sublayer to perform multimodal fusion.

\subsection{Task Definition}
Suppose our dataset has $N$ data-points, we can represent the $i$-th data as $\{U_{j}^i,Y_{j}^i\}$, $U_{j}^i = (X_{j}^i,V_{j}^i)$, where $i \in\left\{ 1,2,...,N\right\}$, $j \in\left\{ 1,2,...,N_i\right\}$, which is a collection of $\{utterance,label\}$ pairs, $N$ denotes the number of conversations, and $N_i$ denotes the number of utterances in the $i$-th conversation. Each utterance consists of two modalities: text (X), video (V). We align the visual features with their corresponding tokens in the text modality. Therefore, both two modalities have the same length. Given an utterance, our task is to predict its emotion label. The objective of the task is to maximize the following function: 
\begin{equation}
 \Theta = \prod _{i=1}^{N} \prod _{j=1}^{N_i} p(Y_{j}^i|U_{j}^i,U_{j-1}^i,...,U_{1}^i;\theta)
 \label{objectfunction}
\end{equation}
where $U_{j-1}^i, ..., U_{1}^i$ denote contextual utterances and $\theta$ denotes the model parameters set. We denote the number of contextual utterances as $M$.


\subsection{Bimodal Encoder Layer}
We extract textual and visual features via the bimodal encoder respectively. For text representation, we use a debiased word embedding layer to convert each token $t$ in $X^{i}$ into a vector representation $\vec{t} \in {\mathbb{R}}^{d}$, where $d$ denotes the size of word embedding. Moreover, the debiased GloVe embeddings (through our debiasing methods presented in Sec.~3.1) are used for initialization in the word embedding layer. Let 

\begin{equation}
  \vec{t} = \mathrm{Embed(t)}
 \label{embedding}
\end{equation}
as described in the previous part, we use a sentiment embedding layer to convert each token $t$ in the utterance into a corresponding sentiment features score $\vec{S_i}$ as an additional information source vector. The resulting textual embeddings are fed into the Transformer encoders to further refine textual representation.

For the visual representation, each input video clip is scaled to $480 \times 360$, and the pre-trained EfficientNet~\cite{tan2019efficientnet} is used to extract the features. The Transformer encoders are used to learn the visual representations.

\subsection{Sentiment Knowledge Attention}
In philosophy and psychology, sentiment and emotion are closely related, corresponding to internal and external human affection~\cite{evans2002emotion}. Sentiment refers to human’s subjective experience and mental attitude, which involves long-term and deep human cognition~\cite{dolan2002emotion}. Therefore, we hypothesise that the sentiment knowledge will help the task of emotion recognition.
Correspondingly, we propose a sentiment knowledge attention mechanism to capture and counterpoise the sentiment representation for each token. Specifically, a gated unit is used to combine the sentiment representation and the original utterance representation.

In our model, we use a commonsense emotion lexicon NRC$\_$VAD~\cite{mohammad2018obtaining} as the sentiment knowledge source. The NRC Valence, Arousal, and Dominance (VAD) lexicon include a list of more than 20,000 English words and their valence, arousal, and dominance scores. 

For a given word and a dimension (V/A/D), the scores range from 0 to 1. 

In general, for each word token $t$ in ${X_j^i}$, we only retrieve its valence values from the NRC$\_$VAD dictionary, which is the `positive-negative' dimension. The final sentiment knowledge representation for each text utterance ${X_j^i}$ is a list of valence scores: [V($t_1$),V($t_2$),...V($t_n$)]. The valence scores of tokens that are not included in NRC$\_$VAD are set to 0.5. The sentiment knowledge representation of each text utterance will be used to enrich the text representation and serve the multi-bias mitigation.
The gate value $g_i$ for each token $x_i$ is calculated as follows:
\begin{equation}
  g_i = \sigma(W_gh_i+b_g)
\label{gate_gi}
\end{equation}
where $h_i$ is the hidden vector of token $x_i$ from the previous $lstm$ layer, $W_g$ is a learnable linear transformation and $b_g$ is the bias. Then the attention output $\vec{T}$ is calculated as a weighted combination of sentiment enriched and original attention scores:
\begin{equation}
  \vec{T_i} = g_i\vec{t_i} + (1-g_i)\vec{S_i}\vec{t_i}    
\label{sentiment_enrich_rep}
\end{equation}

\subsection{Bimodal Cross Attention}
We use a bimodal cross attention layer~\cite{hasan2021humor}, which is a multi-head self-attention mechanism, to learn the joint representation of $U_l$ and $U_v$, $U_l = \vec{T_i}$, where $U_l$ represents sentiment-enriched textual representation and $U_v$ denotes sentiment-enriched visual representation. 

Specifically, we create corresponding sets of queries ($Q_l, Q_v$), keys ($K_l, K_v$), and values ($V_l, V_v$) to learn the interaction between textual and visual modalities ($U_l, U_v$). The modal representation and query set is attached to a multi-head cross attention layer. We also add the normalization layer and residual connections layer after each cross attention layer. Let 
 \begin{equation}
  M_{l,v} = \mathrm{BimodalCrossAttention(U_l,U_v) }  
\label{bimodal_corss_attn}
\end{equation}
\vspace{-2em}
\subsection{Classification}
 The bimodal fusion representation is gained from the bimodal cross attention layer, which is shown in Eq. \ref{bimodal_corss_attn}. We then add a maxpooling layer to extract the most salient features across the time dimension and yield a one-dimensional vector. Let

 \begin{equation}
  M_{l,v} = \mathrm{MaxPooling(M_{l,v})}  
\label{max_pool}
\end{equation}
 \begin{equation}
  P = \mathrm{softmax(M_{l,v}W+b)}
\label{final_representation}
\end{equation}
where $P$ represents the output probability, $W \in \mathbb{R}^{d*l}$ and $b \in \mathbb{R}^{l}$ denote parameters, $l$ denotes the number of classes.

\section{Experiments}

\begin{figure*}[ht]
  \centering
  
    \subfigure[GloVe(gender)]{\includegraphics[width=0.18\textwidth]{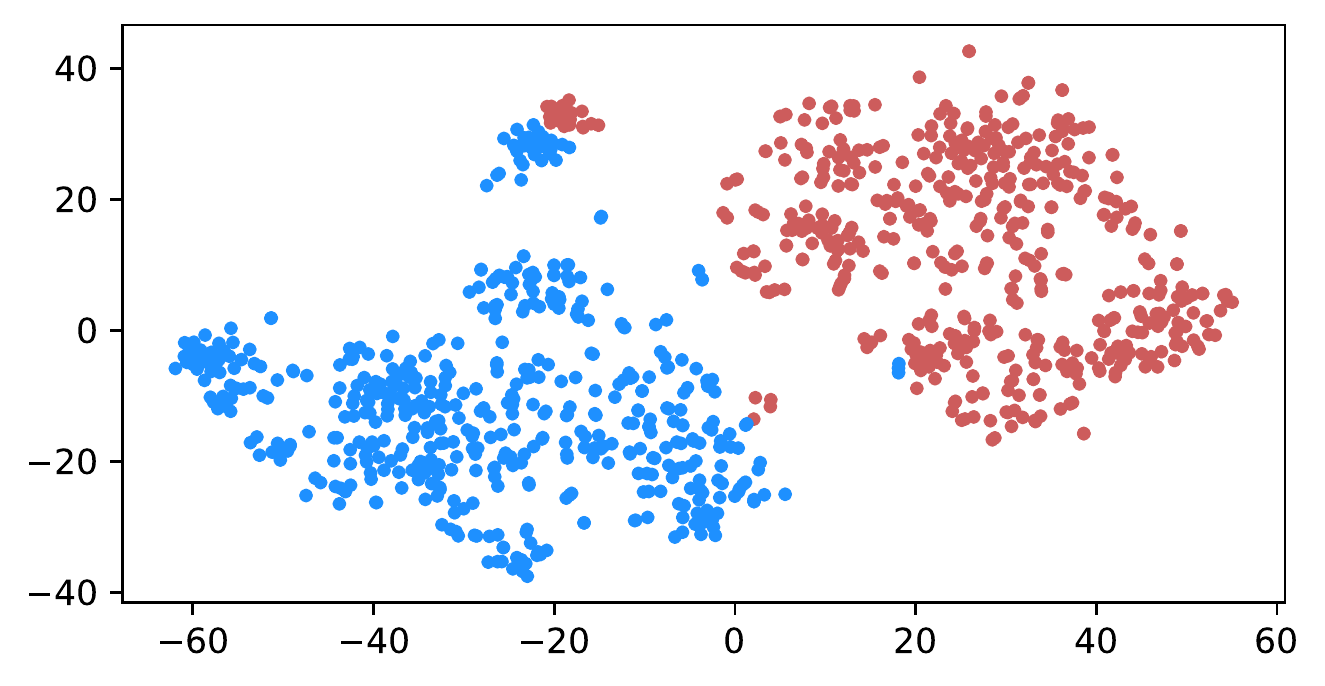}} 
    \subfigure[GloVe(race)]{\includegraphics[width=0.18\textwidth]{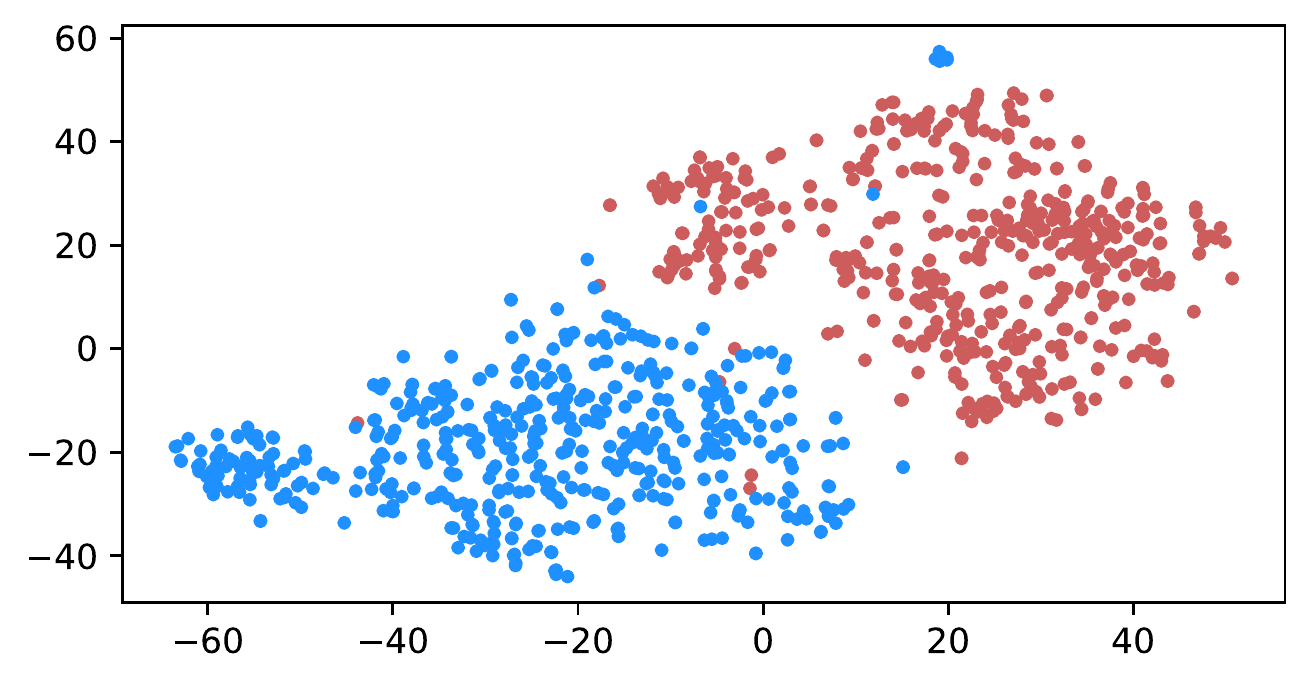}} 
    \subfigure[GloVe(age)]{\includegraphics[width=0.18\textwidth]{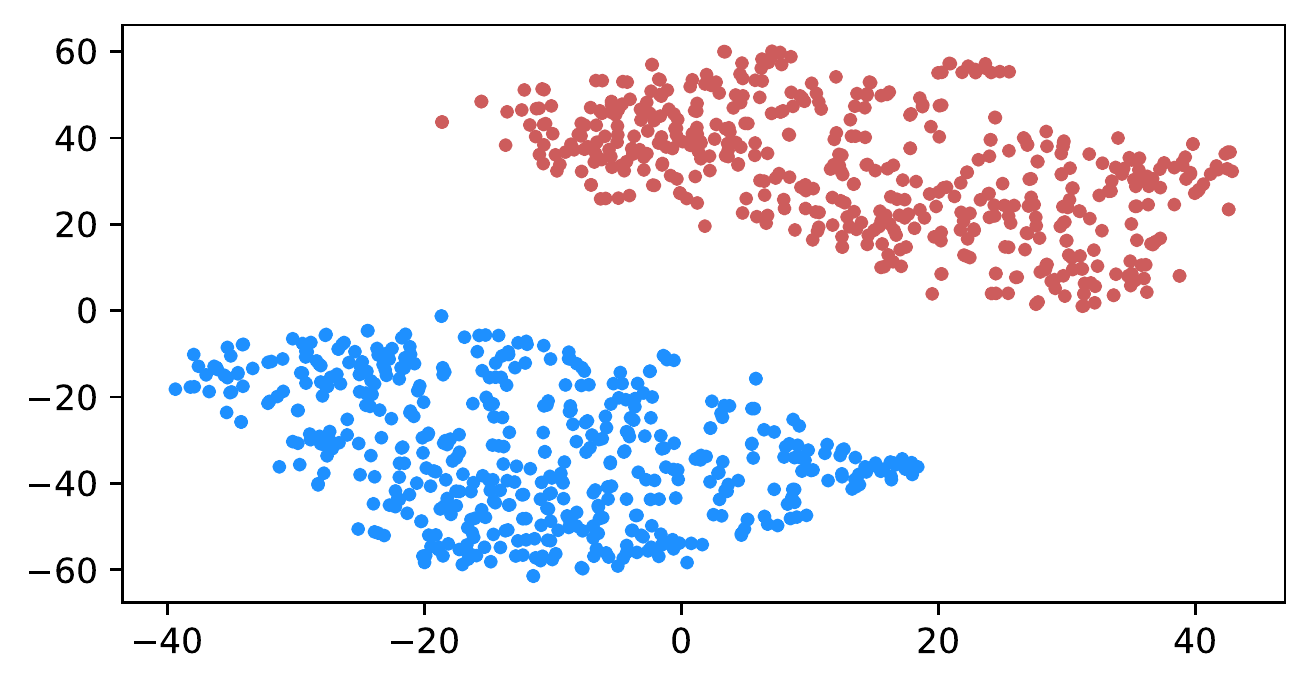}} 
    \subfigure[GloVe(religion)]{\includegraphics[width=0.18\textwidth]{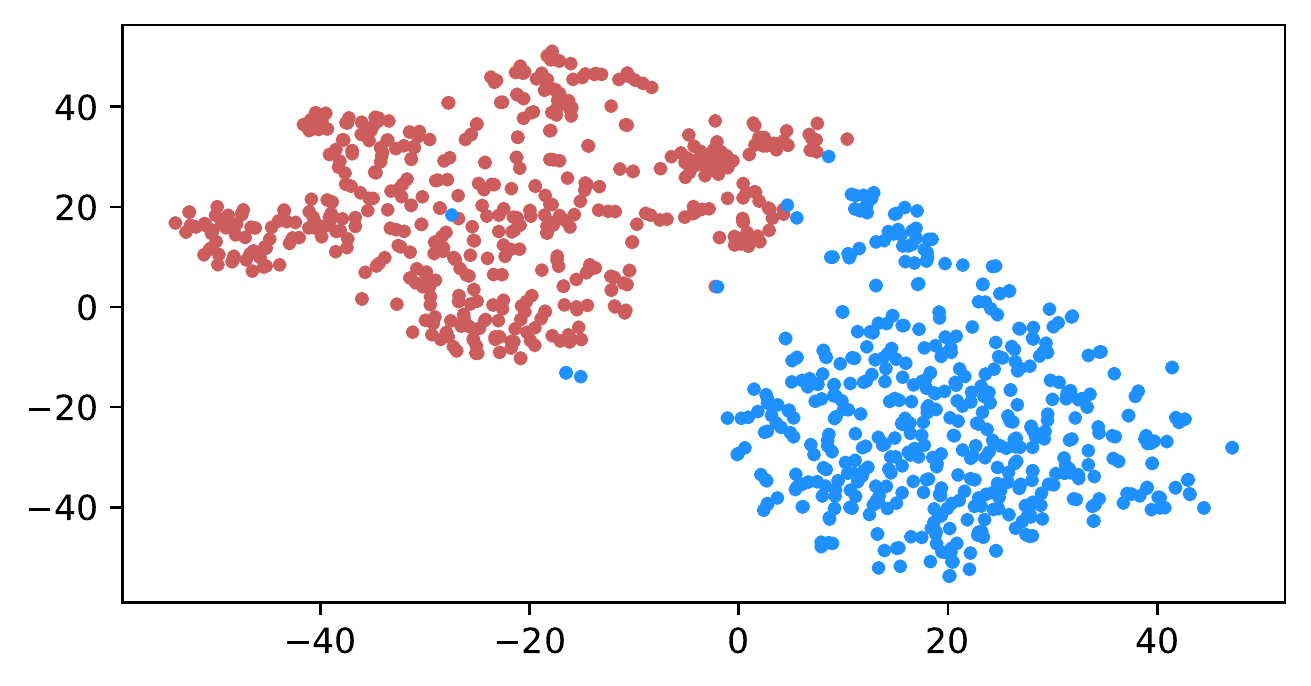}} 
    \subfigure[GloVe(LGBTQ+)]{\includegraphics[width=0.18\textwidth]{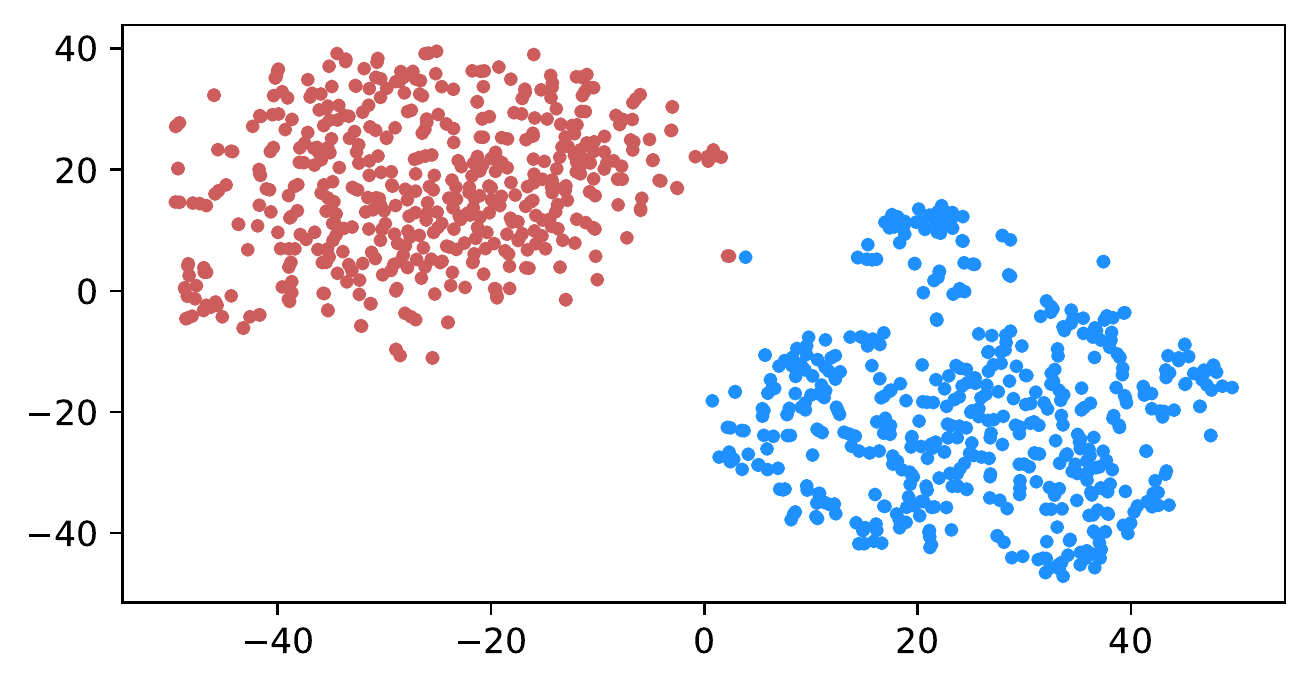}} \\
    \subfigure[Gender-debiased]{\includegraphics[width=0.18\textwidth]{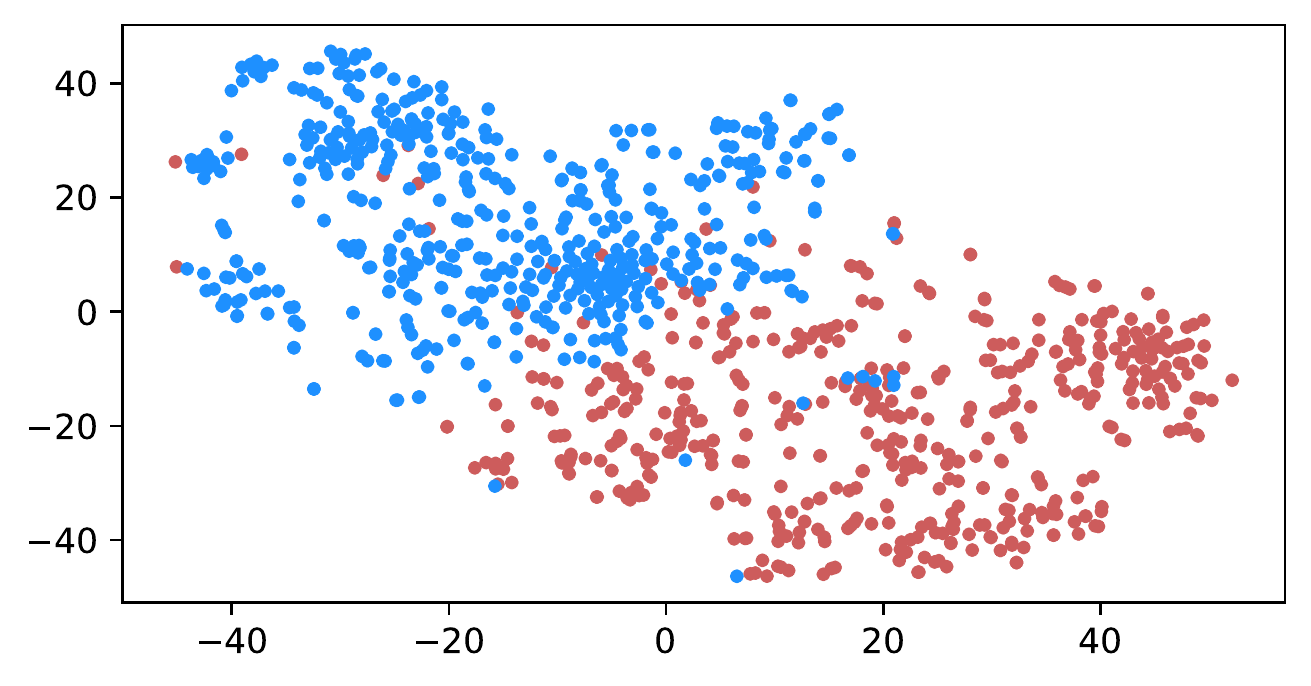}} 
    \subfigure[Race-debiased]{\includegraphics[width=0.18\textwidth]{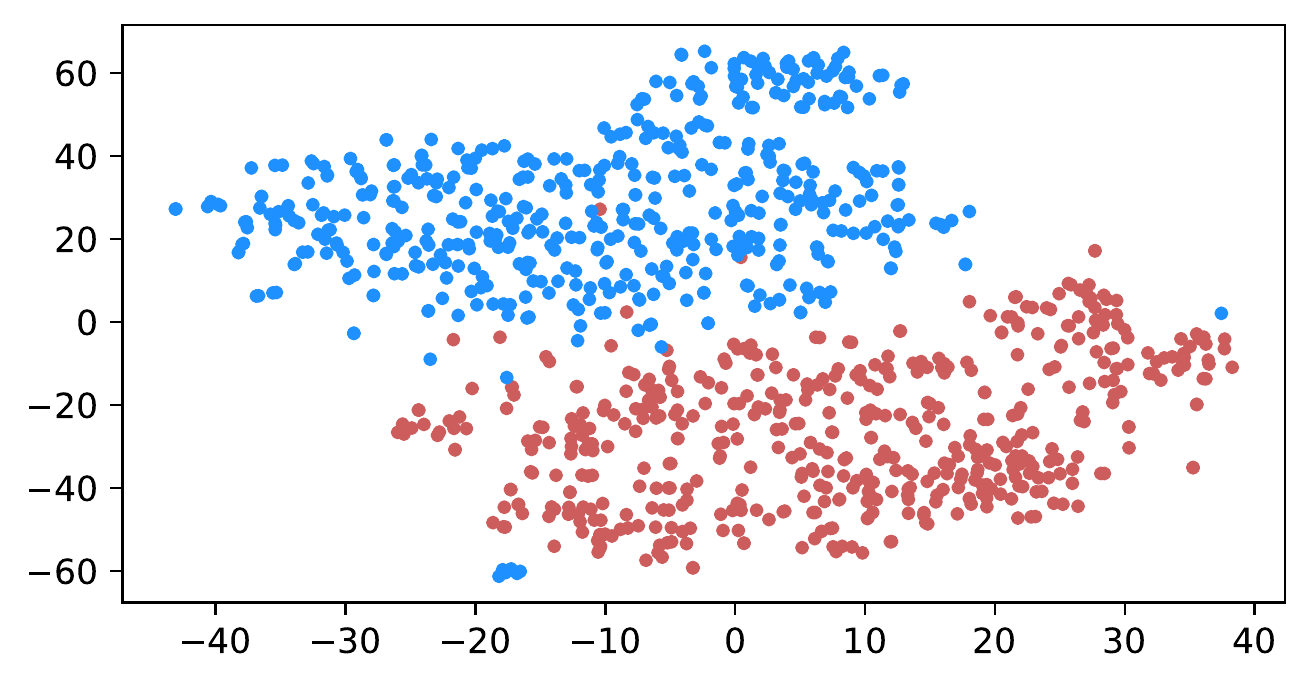}} 
    \subfigure[Age-debiased]{\includegraphics[width=0.18\textwidth]{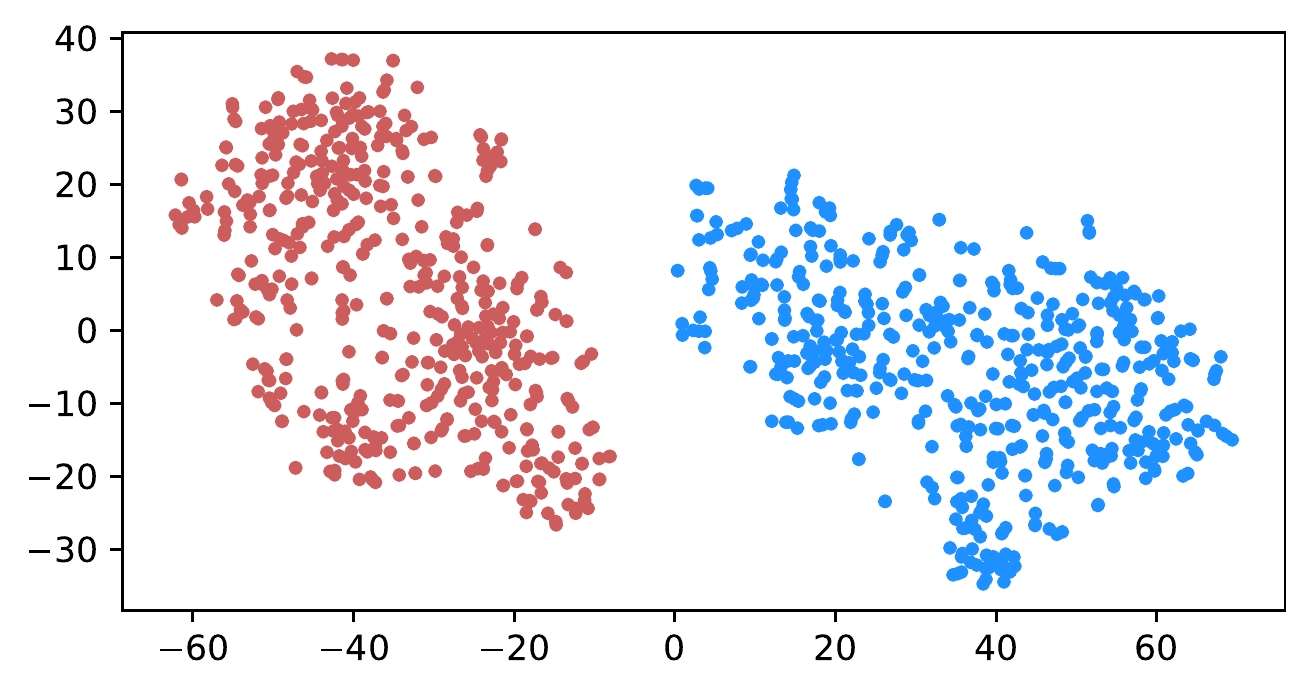}} 
    \subfigure[Religion-debiased]{\includegraphics[width=0.18\textwidth]{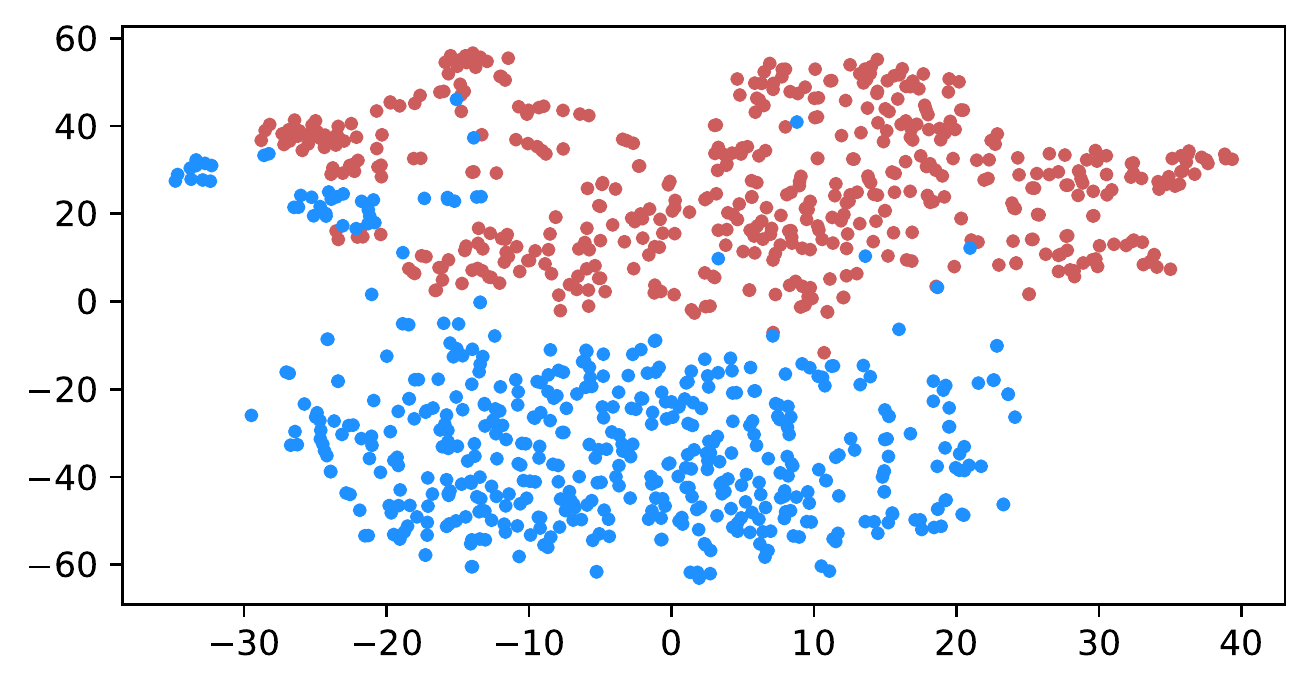}} 
    \subfigure[LGBTQ+-debiased]{\includegraphics[width=0.18\textwidth]{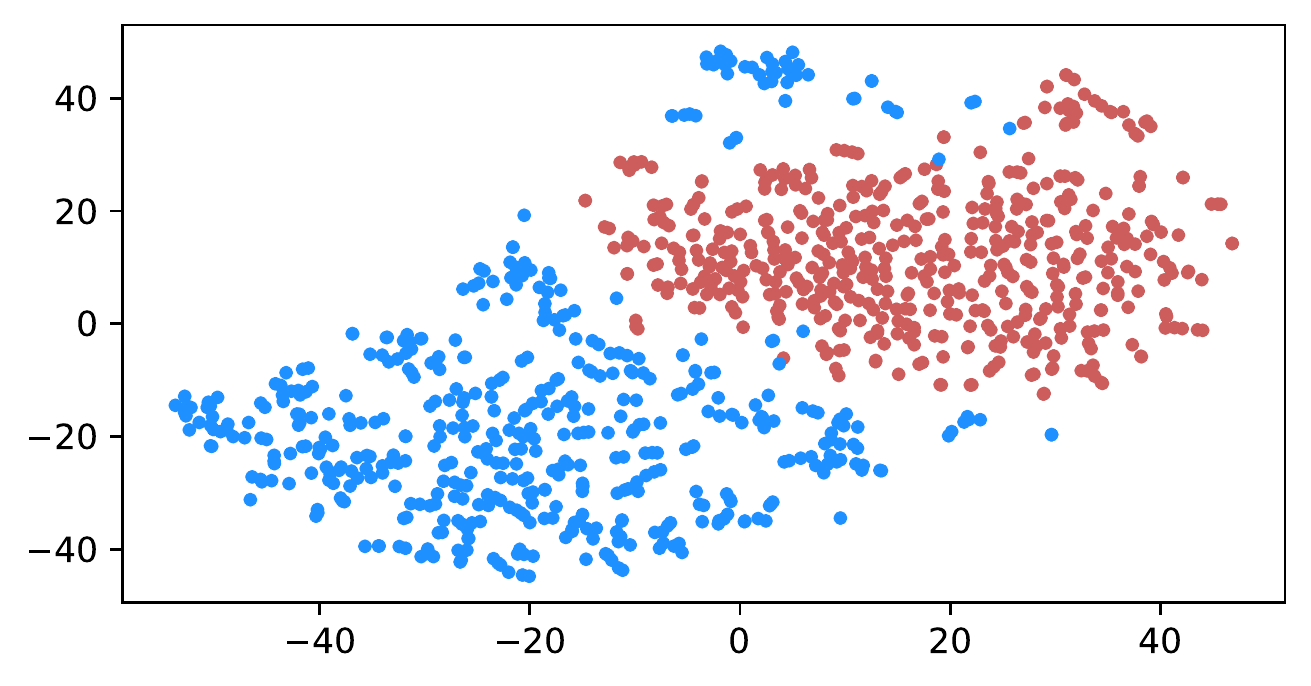}} \\
  \caption{ TSNE visualization of clustering top 500 most biased embeddings (a-e) and their debiased embeddings (f-j).}
  \label{glove-debias-plot}  
\end{figure*}

\subsection{Datasets}

\paragraph{\textbf{IEMOCAP}}~\cite{busso2008iemocap}: A multimodal dataset containing emotional dialogues. Each video contains a single dynamic dialogue, segmented into utterances. The emotion labels of utterances include neutral, happiness, sadness, anger, frustrated, and excited.

\paragraph{\textbf{MELD}}~\cite{poria2019meld}: A dataset of TV show scripts collected from \texttt{Friends}, which is a multimodal emotion classification dataset. The emotion labels of the dataset include happiness, surprise, sadness, anger, disgust, and fear.

Both datasets contain textual, visual, and acoustic information for every utterance. We only focus on the textual and visual modalities in this work. Table \ref{tab1} shows the statistics of the datasets. In all our experiments, 300-dimensional GloVe is leveraged to initialize word embeddings, pre-trained EfficientNet network is used to extract the corresponding feature vectors of images. The dimensionality of hidden states is set to 300. We use adam as an optimizer with a learning rate of 0.0001 and train. The coefficient of L2 regularization is 10\textsuperscript{-5}. The batch size is 64. The network is subjected to regularization in the form of Dropout.

\begin{table}[ht]
\centering
\resizebox{0.36\textwidth}{!}{
\begin{tabular}{ccccc}
\toprule
\multicolumn{2}{c}{Dataset}           & \# dialogues. & \# utterances. \\ 
\midrule
\multirow{3}{*}{\textsc{IEMOCAP}}     & train & 100      & 4810         \\
\cmidrule{2-4}
                                      & dev     &20       &1000      \\
\cmidrule{2-4}
                                      & test   & 31       & 1623          \\ 
\midrule
\multirow{3}{*}{\textsc{MELD}} & train & 1039     & 9989         \\
\cmidrule{2-4}
                            & dev  & 114      & 1109           \\
\cmidrule{2-4}
                            & test   & 280       & 2610           \\
\bottomrule
\end{tabular}}

\caption{The data statistics of \texttt{IEMOCAP} and \texttt{MELD}. }
\label{tab1}  
\end{table}

\subsection{Evaluation Metrics}
\paragraph{\textbf{Debiasing.}}
We use k-Means clustering to verify the effectiveness of the debiasing methods. For each type of bias, we take the top 100/500/1000 of the original GloVe embeddings and 100/300/500 of the visual features by calculating their cosine similarity with the specific bias directions. Then, we cluster them into two groups and compute the alignment accuracy for the bias. To visualize the difference, we applied tSNE projection on word embeddings and the image features.

\paragraph{\textbf{Our Proposed MMKET Model.}}
We evaluate our proposed MMKET model on IEMOCAP and MELD, and adopt the F1 score on the test set as our evaluation metric.

\begin{figure}[ht]
  \centering
    \subfigure[Gender-biased Images]{\includegraphics[width=0.22\textwidth]{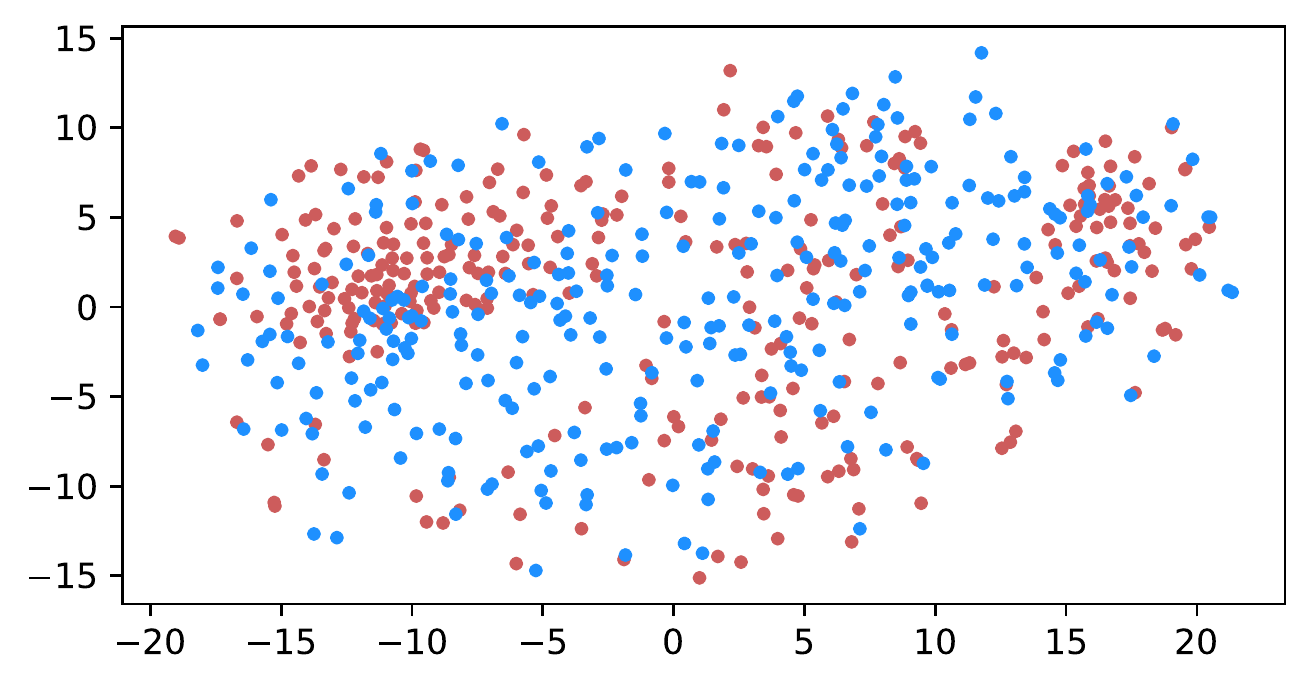}} 
    \subfigure[Age-biased Images]{\includegraphics[width=0.22\textwidth]{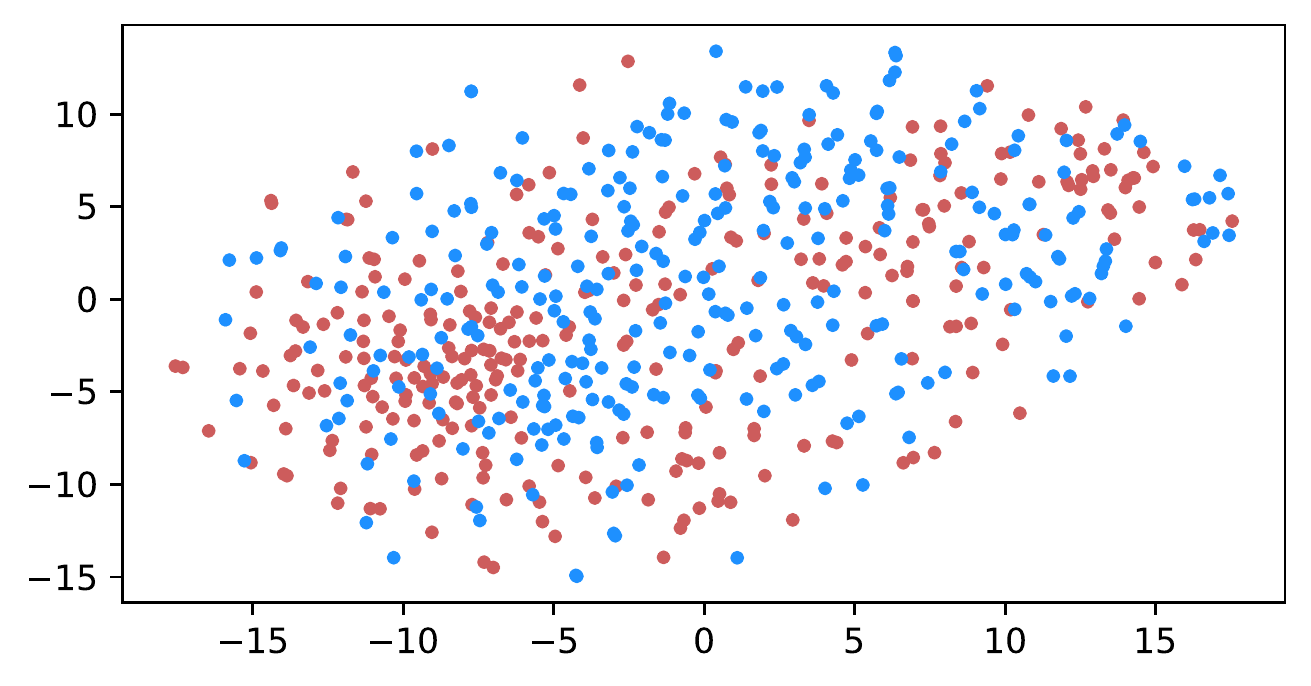}} \\
    \subfigure[Gender-debiased Images]{\includegraphics[width=0.22\textwidth]{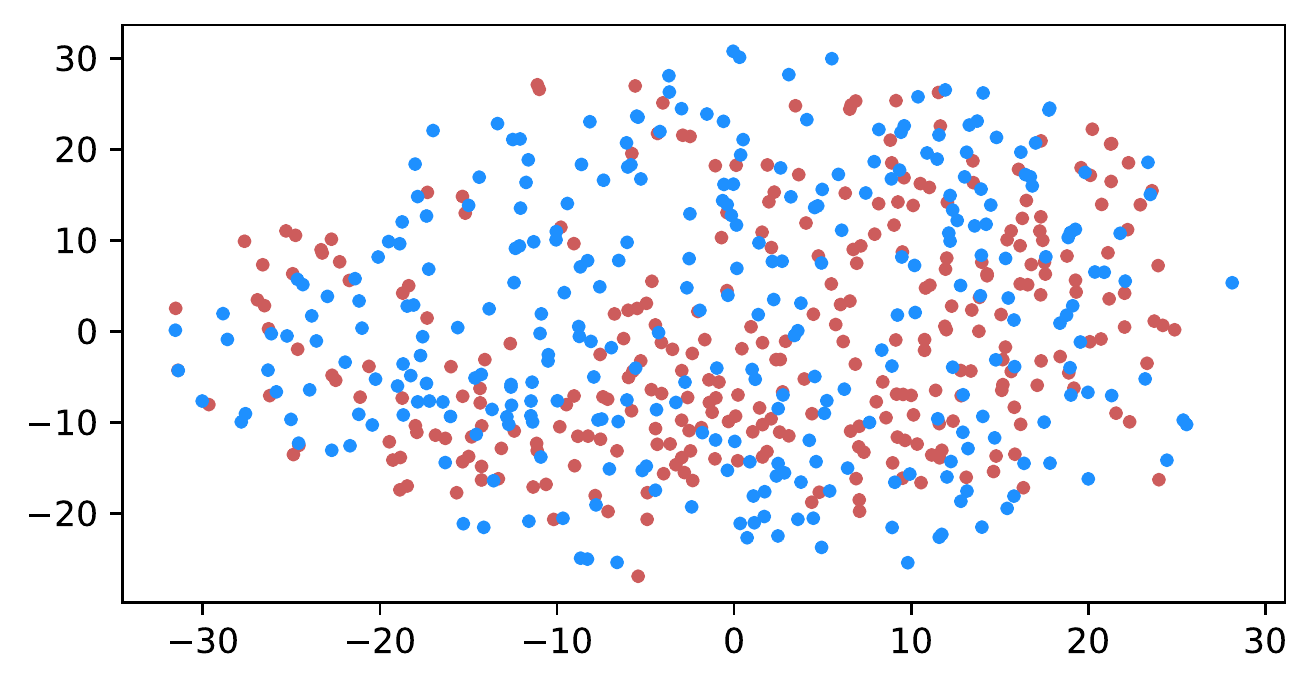}} 
    \subfigure[Age-debiased Images]{\includegraphics[width=0.22\textwidth]{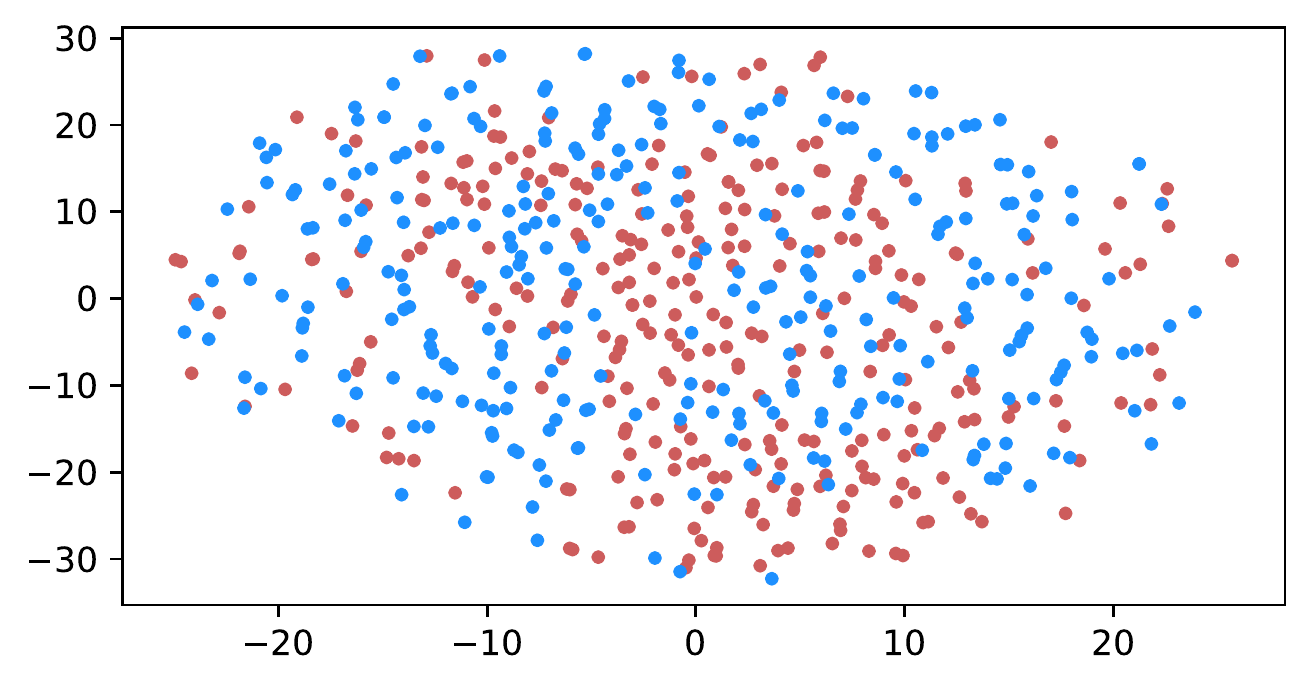}} \\
 
  \caption{ tSNE visualization of top 300 most biased images (a-b) and their debiased counterparts (c-d).}
  \label{figure-debias-plot}  
\end{figure}

\section{Results and Analysis }

\begin{table}[ht]
\centering
 \resizebox{0.42\textwidth}{!}{
\begin{tabular}{cccc}
\toprule
\textbf{Embeddings}     & \textbf{Top100} & \textbf{Top500} & \textbf{Top1000} \\ \hline
GloVe                   & 100.0           & 99.9            & 99.7             \\
Gender-debiased GloVe   & 86.0            & 68.7            & \textbf{55.3}    \\ \hline
GloVe                   & 100.0           & 100.0           & 99.3             \\
Age-debiased GloVe      & 100.0           & 99.2            & \textbf{98.9}    \\ \hline
GloVe                   & 86.5            & 75.3            & 54.5             \\
Race-debiased GloVe     & 86.5            & 75.1            & \textbf{54.4}    \\ \hline
GloVe                   & 99.5            & 95.7            & 96.6             \\
Religion-debiased GloVe & 97.0            & 86.8            & \textbf{81.6}    \\ \hline
GloVe                   & 100.0           & 99.7            & 99.3             \\
LGBTQ+-debiased GloVe   & 94.5            & \textbf{90.7}   & 91.1             \\ 
\bottomrule
\end{tabular}}
\caption{ K-Means clustering accuracy (\%) of top 100/500/1000 biased words.}
\label{K-Means_Result} 
\end{table}
\begin{table}[ht]
\centering
\resizebox{0.3\textwidth}{!}{
\begin{tabular}{ccc}
\hline
\textbf{Mitigated Bias} & \textbf{IEMOCAP} & \textbf{MELD} \\ \hline
None                    & 57.11             & 53.93          \\
Gender                  & 56.22             & 53.22          \\
Race                    & 56.63             & 53.41          \\
Age                     & 56.76             & 53.69          \\
Religion                & 56.09             & 53.14          \\
LGBTQ+                  & 56.89             & 53.20          \\ \hline
\textbf{5 Biases}       & \textbf{55.85}    & \textbf{52.86}         \\ \hline
\end{tabular}}
\caption{\textbf{$T + V$} results (\%) on \textsc{IEMOCAP} and \textsc{MELD} dataset, which is mitigated the specific bias.}
\label{text modal on IEMOCAP and MELD}
\end{table}

\subsection{Effects of Debiasing}
\paragraph{\textbf{Mitigating Multiple Biases in GloVe.}}
Table \ref{K-Means_Result} shows the result of K-Means clustering on the original GloVe and the debiased ones. Lower accuracy means fewer bias cues can be learned. The accuracy appears to decrease after the debiasing operation, suggesting the debias method works effectively in embeddings. More intuitively, in the upper row of Figure \ref{glove-debias-plot}, word embeddings are divided into two clear parts. In the lower row, the two parts have mixed up, though different biases have varied effects. Among the five proposed biases, gender and religious bias were mitigated most. LGBTQ+ bias also reduced, while racial and age bias did not decrease significantly. We speculate that the racial bias is more implicit in textual data given that the accuracy of the original GloVe is already close to 50. 
As for the age bias, we consider the bias words like ``old'' are widely used as unbiased meanings, i.e. ``an old tree'', ``a seven-year-old boy'', which decreased the effect of debiasing. Mitigating the racial and age bias will be left to our future work. 

\paragraph{\textbf{Mitigating Multiple Biases in Visual Representation.}}
Table \ref{K-Means_Result_image} shows the clustering result of biased images. As shown in Figure \ref{figure-debias-plot}, the visual representation of images from IMDB-WIKI are projected into a 2D space. Our proposed debiasing methods mix up the images to a noticeable extent, indicating that gender and age bias are mitigated in image representation.
\begin{table}[ht]
\centering
 \resizebox{0.42\textwidth}{!}{
\begin{tabular}{cccc}
\toprule
\textbf{ Visual Representation}     & \textbf{Top100} & \textbf{Top300} & \textbf{Top500} \\ \hline
Gender-biased Images            &   74.3         &   70.8         &    64.5     \\
Gender-debiased Images   &    61.0         &     59.3       & \textbf{53.3}    \\ \hline
Age-biased Images               &   67.2        &  62.3          &     59.6         \\
Age-debiased Images     &   60.7         &   53.8          & \textbf{52.5}    \\ \hline

\end{tabular}}
\caption{ K-Means clustering accuracy (\%) of top 100/300/500 biased images. Lower accuracy means less bias cues. }
\label{K-Means_Result_image} 
\end{table}

\subsection{Debiased mERC Results}
We make the first step to explore the role of bias plays in mERC tasks. Human emotions contain prejudice, so removing the bias will decrease the emotion classification accuracy, which can explain the results in Table \ref{text modal on IEMOCAP and MELD} and Table \ref{text+vision modal on IEMOCAP and MELD}. Compared to the single modal results (Table 4), our MMKET model makes full use of the rich information in the Bimodal data and the connection between them, which greatly improves the performance of the algorithm.

\begin{table}[ht]
\centering
\resizebox{0.34\textwidth}{!}{
\begin{tabular}{@{}cccc@{}}
\toprule
\multicolumn{2}{c}{\textbf{Mitigated Bias}}          & \multirow{2}{*}{\textbf{IEMOCAP}} & \multirow{2}{*}{\textbf{MELD}} \\ \cmidrule(r){1-2}
\multicolumn{1}{c}{\textbf{Text}} & \textbf{Visual} &                                   &                                \\ \midrule
\multirow{2}{*}{None}              & None            & 58.29                              & 56.35                           \\
                                   & Gender\&Age         & 57.61                              & 55.09
                                   
                                   \\\midrule
\multirow{2}{*}{5 Biases}          & None            & 57.56                              & 55.64                           \\
                                   & Gender\&Age          & 56.23                              & 54.25   
                                 
                                   \\\bottomrule
\end{tabular}
}
\caption{\textbf{$T + V$} results (\%) of mERC on \textsc{IEMOCAP MELD} dataset, by mitigating different types of bias.}
\label{text+vision modal on IEMOCAP and MELD}
\end{table}

\begin{table}[ht]
\centering
\resizebox{0.48\textwidth}{!}{
\begin{tabular}{cllll}
\hline
\textbf{Dataset} & 0       & 0.3                 & 0.5        & 0.7       \\ \hline
MELD             & 55.93   & $\textbf{56.35}_{+0.42}$  & $56.24_{+0.31}$  & $56.09_{+0.16}$    \\
IEMOCAP          & 57.60   & $\textbf{58.29}_{+0.69}$  & $57.96_{+0.36}$  & $57.22_{-0.37}$   \\ \hline
Debiased-MELD    & 54.14   & $\textbf{54.42}_{+0.28}$  & $54.31_{+0.17}$  & $54.25_{+0.11}$   \\
Debiased-IEMOCAP & 56.37   & $\textbf{56.57}_{+0.20}$ & $56.40_{+0.03}$  & $56.38_{+0.01}$   \\ \hline
\end{tabular}
}
\caption{Analysis of the weight of sentiment knowledge from 0 to 0.7.}
\label{Analysis_of_the_weight_of_sentiment_knowledge}
\end{table}

\subsection{Ablation Studies}
To further investigate how the sentiment knowledge affects the debias method and mERC, we conduct extensive ablation experiments with the weight of sentiment knowledge of different values, whose results are included in Table \ref{Analysis_of_the_weight_of_sentiment_knowledge}. The sentiment knowledge improves model performance significantly, but less on the debiased model. One possible reason is that biases themselves imply the emotions of humans, so mitigating biases will reduce the effect of sentiment knowledge.

\section{Conclusion}
In this work, we extend the types of bias in the embedding level (e.g., gender, age, race, religion, and LGBTQ+) and innovatively propose the Projection Debias to mitigate gender and age bias in visual representation. We also present a Multibias-mitigated and Sentiment Knowledge Enriched Transformer (MMKET), taking the first step to explore how the debiasing operation affects the algorithm in multimodal emotion recognition in conversation (mERC). 
We conduct extensive experiments to show the effectiveness of the proposed model and prove that debias operation and sentiment knowledge has a great impact on the classification performance for the task of mERC. Due to the difference of the biases, the effect of debiasing also varies, which requires further research. Our model also has a few limitations. For example, we only select to mitigate two typical visual biases, while other typles of bias are ignored. Such efforts will be left to our future work.
We hope our study will benefit the development of bias mitigation in mERC and other emotion studies.

\section*{Acknowledgements}
This research was supported in part by Natural Science Foundation of Beijing (grant number:  4222036) and Huawei Technologies (grant number: TC20201228005). This work was supported by National Science Foundation of China under grant No. 62006212, the fund of State Key Lab. for Novel Software Technology in Nanjing University (grant No. KFKT2021B41), and the Industrial Science and Technology Research Project of Henan Province (grant No. 222102210031).

\bibliography{acl_latex}

\begin{thebibliography}{28}
\expandafter\ifx\csname natexlab\endcsname\relax\def\natexlab#1{#1}\fi

\bibitem[{Bolukbasi et~al.(2016)Bolukbasi, Chang, Zou, Saligrama, and
  Kalai}]{bolukbasi2016man}
Tolga Bolukbasi, Kai-Wei Chang, James~Y Zou, Venkatesh Saligrama, and Adam~T
  Kalai. 2016.
\newblock Man is to computer programmer as woman is to homemaker? debiasing
  word embeddings.
\newblock \emph{Advances in neural information processing systems}.

\bibitem[{Buolamwini and Gebru(2018)}]{buolamwini2018gender}
Joy Buolamwini and Timnit Gebru. 2018.
\newblock Gender shades: Intersectional accuracy disparities in commercial
  gender classification.
\newblock In \emph{Conference on fairness, accountability and transparency}.

\bibitem[{Busso et~al.(2008)Busso, Bulut, Lee, Kazemzadeh, Mower, Kim, Chang,
  Lee, and Narayanan}]{busso2008iemocap}
Carlos Busso, Murtaza Bulut, Chi-Chun Lee, Abe Kazemzadeh, Emily Mower, Samuel
  Kim, Jeannette~N Chang, Sungbok Lee, and Shrikanth~S Narayanan. 2008.
\newblock Iemocap: Interactive emotional dyadic motion capture database.
\newblock \emph{Language resources and evaluation}.

\bibitem[{Caliskan et~al.(2017)Caliskan, Bryson, and
  Narayanan}]{caliskan2017semantics}
Aylin Caliskan, Joanna~J Bryson, and Arvind Narayanan. 2017.
\newblock Semantics derived automatically from language corpora contain
  human-like biases.
\newblock \emph{Science}.

\bibitem[{Dolan(2002)}]{dolan2002emotion}
Raymond~J Dolan. 2002.
\newblock Emotion, cognition, and behavior.
\newblock \emph{science}, 298(5596):1191--1194.

\bibitem[{Drozdowski et~al.(2020)Drozdowski, Rathgeb, Dantcheva, Damer, and
  Busch}]{drozdowski2020demographic}
Pawel Drozdowski, Christian Rathgeb, Antitza Dantcheva, Naser Damer, and
  Christoph Busch. 2020.
\newblock Demographic bias in biometrics: A survey on an emerging challenge.
\newblock \emph{IEEE Transactions on Technology and Society}.

\bibitem[{Evans(2002)}]{evans2002emotion}
Dylan Evans. 2002.
\newblock \emph{Emotion: The science of sentiment}.
\newblock Oxford University Press, USA.

\bibitem[{Garg et~al.(2018)Garg, Schiebinger, Jurafsky, and Zou}]{garg2018word}
Nikhil Garg, Londa Schiebinger, Dan Jurafsky, and James Zou. 2018.
\newblock Word embeddings quantify 100 years of gender and ethnic stereotypes.
\newblock \emph{Proceedings of the National Academy of Sciences}.

\bibitem[{Goyal et~al.(2019)Goyal, Khot, Agrawal, Summers-Stay, Batra, and
  Parikh}]{goyal2019making}
Yash Goyal, Tejas Khot, Aishwarya Agrawal, Douglas Summers-Stay, Dhruv Batra,
  and Devi Parikh. 2019.
\newblock Making the v in vqa matter: Elevating the role of image understanding
  in visual question answering.
\newblock \emph{International Journal of Computer Vision}.

\bibitem[{Hasan et~al.(2021)Hasan, Lee, Rahman, Zadeh, Mihalcea, Morency, and
  Hoque}]{hasan2021humor}
Md~Kamrul Hasan, Sangwu Lee, Wasifur Rahman, Amir Zadeh, Rada Mihalcea,
  Louis-Philippe Morency, and Ehsan Hoque. 2021.
\newblock Humor knowledge enriched transformer for understanding multimodal
  humor.

\bibitem[{Kurita et~al.(2019)Kurita, Vyas, Pareek, Black, and
  Tsvetkov}]{kurita2019measuring}
Keita Kurita, Nidhi Vyas, Ayush Pareek, Alan~W Black, and Yulia Tsvetkov. 2019.
\newblock Measuring bias in contextualized word representations.
\newblock In \emph{Proceedings of the First Workshop on Gender Bias in Natural
  Language Processing}.

\bibitem[{May et~al.(2019)May, Wang, Bordia, Bowman, and
  Rudinger}]{may2019measuring}
Chandler May, Alex Wang, Shikha Bordia, Samuel Bowman, and Rachel Rudinger.
  2019.
\newblock On measuring social biases in sentence encoders.
\newblock In \emph{Proc. of ACL}.

\bibitem[{Misra et~al.(2016)Misra, Lawrence~Zitnick, Mitchell, and
  Girshick}]{misra2016seeing}
Ishan Misra, C~Lawrence~Zitnick, Margaret Mitchell, and Ross Girshick. 2016.
\newblock Seeing through the human reporting bias: Visual classifiers from
  noisy human-centric labels.
\newblock In \emph{Proc. of CVPR}, pages 2930--2939.

\bibitem[{Mohammad(2018)}]{mohammad2018obtaining}
Saif Mohammad. 2018.
\newblock Obtaining reliable human ratings of valence, arousal, and dominance
  for 20,000 english words.
\newblock In \emph{Proc. of ACL}.

\bibitem[{Nadeem et~al.(2020)Nadeem, Bethke, and Reddy}]{nadeem2020stereoset}
Moin Nadeem, Anna Bethke, and Siva Reddy. 2020.
\newblock Stereoset: Measuring stereotypical bias in pretrained language
  models.
\newblock \emph{arXiv preprint arXiv:2004.09456}.

\bibitem[{Pennington et~al.(2014)Pennington, Socher, and
  Manning}]{pennington2014glove}
Jeffrey Pennington, Richard Socher, and Christopher~D Manning. 2014.
\newblock Glove: Global vectors for word representation.
\newblock In \emph{Proc. of EMNLP}.

\bibitem[{Plank et~al.(2014)Plank, Hovy, and S{\o}gaard}]{plank2014learning}
Barbara Plank, Dirk Hovy, and Anders S{\o}gaard. 2014.
\newblock Learning part-of-speech taggers with inter-annotator agreement loss.
\newblock In \emph{Proc. of ACL}.

\bibitem[{Poria et~al.(2019)Poria, Hazarika, Majumder, Naik, Cambria, and
  Mihalcea}]{poria2019meld}
Soujanya Poria, Devamanyu Hazarika, Navonil Majumder, Gautam Naik, Erik
  Cambria, and Rada Mihalcea. 2019.
\newblock Meld: A multimodal multi-party dataset for emotion recognition in
  conversations.
\newblock In \emph{Proc. of ACL}.

\bibitem[{Prates et~al.(2019)Prates, Avelar, and Lamb}]{prates2019assessing}
Marcelo~OR Prates, Pedro~H Avelar, and Lu{\'\i}s~C Lamb. 2019.
\newblock Assessing gender bias in machine translation: a case study with
  google translate.
\newblock \emph{Neural Computing and Applications}.

\bibitem[{Rothe et~al.(2018)Rothe, Timofte, and Gool}]{Rothe-IJCV-2018}
Rasmus Rothe, Radu Timofte, and Luc~Van Gool. 2018.
\newblock Deep expectation of real and apparent age from a single image without
  facial landmarks.
\newblock \emph{International Journal of Computer Vision}, 126(2-4):144--157.

\bibitem[{Rudinger et~al.(2018)Rudinger, Naradowsky, Leonard, and
  Van~Durme}]{rudinger2018gender}
Rachel Rudinger, Jason Naradowsky, Brian Leonard, and Benjamin Van~Durme. 2018.
\newblock Gender bias in coreference resolution.
\newblock In \emph{Proc. of ACL}.

\bibitem[{Sap et~al.(2019)Sap, Card, Gabriel, Choi, and Smith}]{sap2019risk}
Maarten Sap, Dallas Card, Saadia Gabriel, Yejin Choi, and Noah~A Smith. 2019.
\newblock The risk of racial bias in hate speech detection.
\newblock In \emph{Proc. of ACL}.

\bibitem[{Splieth{\"o}ver and Wachsmuth(2021)}]{spliethover2021bias}
Maximilian Splieth{\"o}ver and Henning Wachsmuth. 2021.
\newblock Bias silhouette analysis: Towards assessing the quality of bias
  metrics for word embedding models.
\newblock In \emph{Proceedings of the Thirtieth International Joint Conference
  on Artificial Intelligence, IJCAI-21}.

\bibitem[{Srinivasan and Bisk(2021)}]{srinivasan2021worst}
Tejas Srinivasan and Yonatan Bisk. 2021.
\newblock Worst of both worlds: Biases compound in pre-trained
  vision-and-language models.
\newblock \emph{arXiv preprint arXiv:2104.08666}.

\bibitem[{Tan and Le(2019)}]{tan2019efficientnet}
Mingxing Tan and Quoc Le. 2019.
\newblock Efficientnet: Rethinking model scaling for convolutional neural
  networks.
\newblock In \emph{Proc. of ICML}.

\bibitem[{Vaswani et~al.(2017)Vaswani, Shazeer, Parmar, Uszkoreit, Jones,
  Gomez, Kaiser, and Polosukhin}]{vaswani2017attention}
Ashish Vaswani, Noam Shazeer, Niki Parmar, Jakob Uszkoreit, Llion Jones,
  Aidan~N Gomez, {\L}ukasz Kaiser, and Illia Polosukhin. 2017.
\newblock Attention is all you need.
\newblock In \emph{Advances in neural information processing systems}.

\bibitem[{Wang et~al.(2020)Wang, Lin, Rajani, McCann, Ordonez, and
  Xiong}]{wang2020double}
Tianlu Wang, Xi~Victoria Lin, Nazneen~Fatema Rajani, Bryan McCann, Vicente
  Ordonez, and Caiming Xiong. 2020.
\newblock Double-hard debias: Tailoring word embeddings for gender bias
  mitigation.
\newblock In \emph{Proc. of ACL}.

\bibitem[{Webster et~al.(2020)Webster, Wang, Tenney, Beutel, Pitler, Pavlick,
  Chen, Chi, and Petrov}]{webster2020measuring}
Kellie Webster, Xuezhi Wang, Ian Tenney, Alex Beutel, Emily Pitler, Ellie
  Pavlick, Jilin Chen, Ed~Chi, and Slav Petrov. 2020.
\newblock Measuring and reducing gendered correlations in pre-trained models.
\newblock \emph{arXiv preprint arXiv:2010.06032}.

\end{thebibliography}
\bibliographystyle{acl_natbib}
\end{document}